\documentclass[10pt,twocolumn,letterpaper]{article}

\usepackage{cvpr}
\usepackage{times}
\usepackage{epsfig}
\usepackage{graphicx}
\usepackage{amsmath}
\usepackage{amssymb}
\usepackage{eucal,bibspacing}

\usepackage[bf]{caption}

\usepackage{cite}

\usepackage{cs}
\usepackage{mathsymb}
\usepackage{verbatim}
\usepackage{subfigure}
\usepackage{url}
\usepackage{graphicx}
\graphicspath{{./}{./Fig/}{./Figs/}{./Fig/eps/}{./Fig/pdf/}}

\usepackage{algorithm2e}

\let\savedalgorithm\algorithm
\let\savedendalgorithm\endalgorithm
\newenvironment{algorithmic}{%
\savedalgorithm
}{%
\savedendalgorithm
}

\usepackage{algorithm}

\usepackage{amsmath}
\usepackage{url}
\usepackage{multirow}
\usepackage[super]{nth}

\def\cone{{\ding{172}}}
\def\ctwo{{\ding{173}}}
\def\cthree{{\ding{174}}}

\def\bPbar{{\bar{\bP}}}

\def\bxi{{\bx_i}}
\def\bxj{{\bx_j}}
\def\bxip{{\bx_i^{+}}}
\def\bxijm{{\bx_{i,j}^{-}}}
\def\mprime{{m^{\prime}}}

\def\bdip{{\bd_{i}^{+}}}
\def\bdjm{{\bd_{j}^{-}}}
\def\bdjbm{{\bd_{(j)}^{-}}}

\def\bPast{{{\boldsymbol P}^{\ast}}}

\def\Integer{\mathbb{Z}^+}

\def\CMCstruct{{\rm CMC$^{\,\rm top}$}\xspace}
\def\CMCtriplet{{\rm CMC$^{\,\rm triplet}$}\xspace}

\renewcommand{\etc}{etc.}

\cvprfinalcopy %

\newcommand{\CScomment}[1]{}

\def\etc{etc.\@\xspace}

\renewcommand{\paragraph}{\textbf}

\begin{document}

\title{Learning to rank in person re-identification with metric ensembles}

\author{
         Sakrapee Paisitkriangkrai,
         Chunhua Shen\thanks{Corresponding author: C. Shen (chhshen@gmail.com).},
         Anton van den Hengel
         \\
         The University of Adelaide, Australia; and Australian Center for Robotic Vision
}

\maketitle

\begin{abstract}

We propose an effective structured learning based approach
to the problem of person re-identification
which outperforms the current state-of-the-art on most benchmark data sets evaluated.
Our framework is built on the basis of multiple low-level hand-crafted
and high-level visual features.
We then formulate two optimization algorithms,
which directly optimize
evaluation measures commonly used in person re-identification, also known as
the Cumulative Matching Characteristic (CMC) curve.
Our new approach is practical to many real-world surveillance applications
as the re-identification performance can be concentrated in the range of most
practical importance.
The combination of these factors leads to a person re-identification system
which outperforms most existing algorithms.
More importantly, we advance state-of-the-art results on person re-identification
by improving the rank-$1$ recognition rates from $40\%$ to $50\%$ on the
iLIDS benchmark, $16\%$ to $18\%$ on the PRID2011 benchmark,
$43\%$ to $46\%$ on the VIPeR benchmark, $34\%$ to $53\%$ on the
CUHK01 benchmark and $21\%$ to $62\%$ on the CUHK03 benchmark.

\end{abstract}

\tableofcontents

\section{Introduction}

The task of person re-identification (re-id) is to match pedestrian images
observed from multiple cameras.
It has recently gained popularity in research community due to its
several important applications in video surveillance.
An automated re-id system could save a lot of human labour in
exhaustively searching for a person of interest from
a large amount of video sequences.

Despite several years of research in the computer vision community,
person re-id is still a very challenging task and
remains unsolved due to (a) large variation in visual appearance
(person's appearance often undergoes large variations
across different camera views);
(b) significant changes in human poses
at the time the image was captured;
(c) large amount of illumination changes and
(d) background clutter and occlusions.
Moreover the problem becomes increasingly difficult when
persons share similar appearance, \eg,
people wearing similar
clothing style with similar color.

To address these challenges, existing research
on this topic
has concentrated on the development
of sophisticated and robust features to describe the visual appearance
under significant changes.
However the system that relies heavily on one specific type
of visual cues, \eg, color, texture or shape,
would not be practical and powerful enough to discriminate individuals with
similar visual appearance.
Existing studies have tried to address the above problem
by seeking a combination of robust and distinctive feature
representation of person's appearance,
ranging from color histogram \cite{Gray2008Viewpoint},
spatial co-occurrence representation \cite{Wang2007Shape},
LBP \cite{Xiong2014Person},
color SIFT \cite{Zhao2013Unsupervised}, \etc

One simple approach to exploit multiple visual features
is to build an ensemble of distance functions,
in which each distance function is learned using
a single feature
and the final distance is calculated from a weighted
sum of these distance functions
\cite{Farenzena2010Person,Xiong2014Person,Zhao2013Unsupervised}.
However existing works on person re-id often pre-define
these weights,
which need to be re-estimated
beforehand for different data sets.
Since different re-id benchmark data sets can have very different
characteristics, \ie, variation in view angle, lighting and occlusion,
combining multiple distance functions using pre-determined
weights is undesirable as highly discriminative features in one environment
might become irrelevant in another environment.

In this paper, we introduce effective approaches
to learn weights of these distance functions.
The first approach optimizes the relative distance using the triplet information
and the second approach maximizes the average
rank-$k$ recognition rate, in which $k$ is chosen to be small,
\eg, $k < 10$.
Setting the value of $k$ to be small is
crucial for many real-world applications since most surveillance
operators typically inspect only the first ten or twenty items retrieved.

The main contributions of this paper are twofold:
1)
We propose two principled approaches to build an ensemble of
person re-id algorithms.
The first approach aims at
maximizing the relative distance between images of
different individuals and images of the same individual
such that the CMC curve approaches one with a minimal
number of returned candidates.
The second approach directly optimizes the
probability that any of these top $k$ matches
are correct using structured learning.
Our ensemble-based approaches are highly flexible and can be combined with linear
and non-linear metrics.
2)
Extensive experiments are carried out to demonstrate that by building
an ensemble of person re-id algorithms learned from different
visual features, notable improvement on rank-$1$
recognition rate can be obtained.
Experimental results show that
our approach achieves the state-of-the-art performance on most person
re-id benchmark data sets evaluated.
In addition, our ensemble approach is complementary to any existing distance learning methods.

\paragraph{Related work}
Existing person re-id systems consist of two major
components: feature representation and metric learning.
In feature representation, robust and discriminative features are
constructed such that they can be used to describe the appearance
of the same individual across different camera views
under various changes and conditions \cite{Bazzani2012Multiple,
Cheng2011Custom,Farenzena2010Person,
Gheissari2006Person,Gray2008Viewpoint,
Wang2007Shape,Zhao2013Unsupervised,Zhao2014Learning}.
We briefly discuss some of these work below.
More feature representations, which have been
applied in person re-id,
can be found in \cite{Gong2013Person}.

Bazzani \etal represent a person by a global mean
color histogram and recurrent local patterns
through epitomic analysis \cite{Bazzani2012Multiple}.
Farenzena \etal propose the symmetry-driven accumulation
of local features which exploits both symmetry and asymmetry,
and represents each part of a person by a weighted color histogram,
maximally stable color regions and texture information \cite{Farenzena2010Person}.
Gray and Tao introduce an ensemble of local features which combines
three color channels with $19$ texture channels \cite{Gray2008Viewpoint}.
Schwartz and Davis propose a discriminative appearance based model using partial least squares,
in which multiple visual features: texture, gradient and color features
are combined \cite{Schwartz2009Learning}.
Zhao \etal propose dcolorSIFT which combines SIFT features with color histogram.
The same authors also propose mid-level filters for person re-identification
by exploring the partial area under the ROC curve (pAUC) score \cite{Zhao2014Learning}.

A large number of metric learning and ranking algorithms
have been proposed \cite{Chopra2005Learning,Davis2007Information,
Frome2007Learning,Kedem2012Nonlinear,Kostinger2012Large,Weinberger2006Distance,
Weinberger2008Fast,Wu2011Optimizing,Xiong2014Person}.
Many of these have been applied to the problem of
person re-id.
We briefly review some of these algorithms.
Interested readers should see \cite{Yang2006Distance}.
Chopra \etal propose an algorithm to learn a similarity metric
from data \cite{Chopra2005Learning}.
The authors train a convolutional network that maps input images into a target
space such that the $\ell_1$-norm in the target space approximate
the semantic distance in the image space.
Gray and Tao use AdaBoost to select discriminative features \cite{Gray2008Viewpoint}.
Koestinger \etal propose the large-scale metric learning
from equivalence constraint
which considers a log likelihood ratio test of two
Gaussian distributions \cite{Kostinger2012Large}.
Li \etal propose the learning of locally adaptive decision functions,
which can be viewed as a joint model of a distance metric
and a locally adapted thresholding rule \cite{Li2013Learning}.
Li \etal propose a filter pairing neural network to learn
visual features for the re-identification task from image data \cite{Li2014Deep}.
Pedagadi \etal combine color histogram with supervised
Local Fisher Discriminant Analysis \cite{Pedagadi2013Local}.
Prosser \etal use pairs of similar and dissimilar images and train the ensemble RankSVM
such that the true match gets the highest rank \cite{Prosser2010Person}.
Weinberger \etal propose the large margin nearest neighbour (LMNN) algorithm to learn the
Mahalanobis distance metric, which improves the k-nearest neighbour classification
\cite{Weinberger2006Distance}.
LMNN is later applied to a task of person re-identification in \cite{Hirzer2012Person}.
Wu \etal applies the Metric Learning to Rank (MLR) method of \cite{McFee2010Metric} to person
re-id \cite{Wu2011Optimizing}.

Although a large number of existing algorithms have
exploited state-of-the-art visual features and
advanced metric learning algorithms,
we observe that the best obtained overall performance on
commonly evaluated person re-id benchmarks, \eg, iLIDS and VIPeR,
is still far from the performance needed for most real-world
surveillance applications.

\paragraph{Notation}
Bold lower-case letters, \eg, $\bw$, denote column vectors and
bold upper-case letters, \eg, $\bP$, denote matrices.
We assume that the provided training data is
for the task of single-shot person re-identification,
\ie, there exist only two images of the same person --
one image taken from camera view A and another image taken from camera view B.
We represent a set of training samples by $\left\{ (\bxi, \bxip) \right\}_{i=1}^m$
where $\bx_i \in {\Real}^D$ represents
a training example from one camera (\ie, camera view A),
and $\bxip$ is the corresponding image of the same person
from a different camera (\ie, camera view B).
Here $m$ is the number of persons in the training data.
From the given training data, we can generate
a set of triplets for each sample $\bxi$ as
$\left\{(\bxi, \bxip, \bxijm) \right\}$
for $i = 1,\cdots,m$ and $i \neq j$.
Here we introduce $\bxijm \in \cX_i^{-}$ where $\cX_i^{-}$ denotes
a subset of images of persons with a different identity to $\bxi$
from camera view B.
We also assume that there exist a set of distance functions
$d_t(\cdot,\cdot)$ which calculate the distance between
two given inputs.
Our goal is to learn a weighted distance function:
$d(\cdot,\cdot) = \sum_{t=1}^T w_t d_t(\cdot,\cdot)$,
such that
the distance between $\bxi$ (taken from camera view A) and $\bxip$
(taken from camera view B)
is smaller than the distance between
$\bxi$ and any $\bxijm$ (taken from camera view B).
The better the distance function, the faster the
cumulative matching characteristic (CMC)
curve approaches one.

\section{Our Approach}
In this section, we propose two approaches that can learn
an ensemble of base metrics.
We then discuss base metrics and visual features
that will be used in our experiment.
\subsection{Ensemble of base metrics}
\label{sec:ensemble}
The most commonly used performance measure for evaluating
person re-id is known as a cumulative
matching characteristic (CMC) curve \cite{Gray2007Evaluating},
which is analogous to the ROC curve in detection problems.
The CMC curve represents results of an identification task
by plotting the probability of correct identification (y-axis)
against the number of candidates returned (x-axis).
The faster the CMC curve approaches one, the better the person re-id algorithm.
Since a better rank-$1$ recognition rate is often preferred \cite{Zhao2014Learning},
our aim is to improve the recognition rate
among the $k$ best candidates, \eg, $k < 20$, which is crucial for many
real-world surveillance applications.
Note that, in practice, the system that achieves the best
recognition rate when $k$ is large (\eg, $k > 100$)
is of little interest since most users inspect or
consider only the first ten or twenty returned candidates.

In this section, we propose two different approaches
which learn an ensemble of base metrics (discussed in the next section).
The first approach, \CMCtriplet, aims at minimizing the number of returned list of candidates
in order to achieve a perfect identification,
\ie, minimizing $k$ such that the rank-$k$ recognition rate is equal to one.
The second approach, \CMCstruct, optimizes the probability that any of these
$k$ best matches are correct.

\subsubsection{Relative distance based approach (\CMCtriplet)}
In order to minimize $k$ such that the rank-$k$ recognition rate
is equal to $100\%$,
we consider learning an ensemble of distance
functions based on relative comparison of triplets \cite{Schultz2004Learning}.
Given a set of triplets
$\left\{(\bxi, \bxip, \bxijm) \right\}_{i,j}$,
in which $\bxi$ is taken from camera view A and
$\{ \bxip, \bxijm \}$ are taken from camera view B,
the basic idea is to learn
a distance function such that
images of the same individual are closer than any images of
different individuals, \ie, $\bxi$ is closer to $\bxip$  than any $\bxijm$.
For a triplet $\left\{ (\bxi, \bxip, \bxijm) \right\}_{i,j}$, the following condition
must hold $d(\bxi,\bxijm) > d(\bxi,\bxip), \forall j, i \neq j$.
Following the large margin framework with the hinge loss, the condition
$d(\bxi,\bxijm) \geq 1 + d(\bxi,\bxip)$ should be satisfied.
This condition means that the distance between two images of
different individuals should be
larger by at least a unit than
the distance between two images of the
same individual.
Since the above condition cannot be satisfied by all triplets,
we introduce a slack variable to enable soft margin.
By generalizing the above idea to the entire training set,
the primal problem that we want to optimize can be written as,
\begin{align}
    \label{EQ:svm}
        \min_{ \bw, \bslack }   \;
        &
         \frac{1}{2} \| \bw  \|_{2}^{2} + \nu \, \frac{1}{m(m-1)} \;
                                \sum_{i=1}^m \sum_{j=1}^{m-1} \xi_{ij}   \\ \notag
        \st \; &
        \bw^\T ( \bdjm - \bdip ) \geq 1 - \xi_{ij}, \forall \left\{i,j\right\}, i \neq j ;
            \\ \notag
        \; &
        \bw \geq 0; \; \bslack \geq 0.
\end{align}
Here $\nu > 0$ is the regularization parameter and
$\bdjm$ = $[d_1(\bxi,\bxijm),$ $\cdots,$ $d_t(\bxi,\bxijm)]$,
$\bdip$ = $[d_1(\bxi,\bxip),$ $\cdots,$ $d_t(\bxi,\bxip)]$ and
$\{d_1(\cdot,\cdot)$,$\cdots$,$d_t(\cdot,\cdot)\}$ represent
a set of base metrics.
Note that we introduce the regularization term $\| \bw \|_{2}^2$ to avoid the trivial solution
of arbitrarily large $\bw$.

We point out here that any smooth convex loss function can also be applied.
Suppose $\lambda(\cdot)$ is a smooth convex function defined in $\Real$
and $\omega(\cdot)$ is any regularization function.
The above optimization problem which enforces the relative comparison
of the triplet can also be written as,
\begin{align}
    \label{EQ:eq2}
        \min_{ \bw }   \;
        &
        \omega( \bw ) + \nu \, \sum_{\tau} \lambda( \rho_{\tau} )    \\ \notag
        \st \; &
        \rho_{\tau} = \sum_t w_t d_t(\bxi,\bxijm) - \sum_{t} w_t d_t(\bxi,\bxip), \forall \tau ;
            \\ \notag
        \; &
        \bw \geq 0,
\end{align}
where $\tau$ being the triplet index set.
In this paper, we consider the hinge loss but other convex loss
functions \cite{Teo2007Scalable} can be applied.

Since the number of constraints in \eqref{EQ:svm}
is quadratic in the number of training examples,
directly solving \eqref{EQ:svm} using off-the-shelf
optimization toolboxes can only solve problems with up to
a few thousand training examples.
In the following, we present an equivalent reformulation of \eqref{EQ:svm},
which can be efficiently solved in a linear runtime using
cutting-plane algorithms.
We first reformulate \eqref{EQ:svm} by writing it as:
\begin{align}
    \label{EQ:svm2}
        \min_{ \bw, \xi }   \;
        &
         \frac{1}{2} \| \bw  \|_{2}^{2} + \nu \,  \xi   \\ \notag
        \st \; &
        \frac{1}{m(m-1)} \bw^\T \Bigl[ \sum_{i=1}^m \sum_{j=1}^{m-1}
                ( \bdjm - \bdip ) \Bigr] \geq 1 - \xi,
            \\ \notag
        \; &
        \forall \left\{i,j\right\}, i \neq j ; \; \bw \geq 0; \; \bslack \geq 0.
\end{align}
Note that the new formulation has a single slack variable.
Later on in this section, we show how the cutting-plane method
can be applied to solve \eqref{EQ:svm2}.

\subsubsection{Top recognition at rank-$k$ (\CMCstruct)}
Our previous formulation assumes that, for any triplets,
images belonging to the same individual should be closer
than images belonging to different individuals.
Our second formulation is motivated by the nature of
the problem, in which person re-id users often
browse only the first few retrieved matches.
Hence we propose another approach,
in which the objective is no longer to
minimize $k$ (the number of returned matches
before achieving $100\%$ recognition rate),
but to maximize the correct identification among
the top $k$ best candidates.
Built upon the structured learning
framework \cite{Joachims2005Support, Narasimhan2013Structural},
we optimize the performance measure commonly used in the CMC curve
(recognition rate at rank-$k$) using structured learning.
The difference between our work and \cite{Narasimhan2013Structural}
is that \cite{Narasimhan2013Structural} assumes training samples
consist of $m_{+}$ positive instances and $m_{-}$ negative instances,
while our work assumes that there are $m$ individuals in camera view A
and $m$ individuals in camera view B.
However there exists ranking in both works:
\cite{Narasimhan2013Structural} attempts to rank all positive samples
before a subset of negative samples while
our works attempt
to rank a pair of the same individual above a pair of different individuals.
Both also apply structure learning of \cite{Joachims2005Support}
to solve the optimization problem.

Given the training individual $\bxi$ (from camera view A) and its correct match
$\bxip$ from camera view B,
we can represent
the relative ordering of all matching candidates in camera view B via a
vector $\bp \in \Real^\mprime$, in which
$p_j$ is $0$ if $\bxip$ (from camera view B) is ranked {\em above}
$\bxijm$ (from camera view B) and
$1$ if $\bxip$ is ranked {\em below} $\bxijm$.
Here $\mprime$ is the total number of individuals from camera view B
who has a different identity to $\bxi$.
Since there exists only one image of the same individual in the camera view B,
$\mprime$ is equal to $m-1$ where $m$ is the total number of individuals
in the training set.
We generalize this idea to the entire training set and represent
the relative ordering via a matrix
$\bP \in \left\{ 0, 1 \right\}^{m \times \mprime}$ as follows:
\begin{align}
    \label{EQ:piij}
        p_{ij}  =
            \begin{cases}
                0  \quad& \text{if} \; \bxip \; \text{is ranked above} \; \bxijm  \\
                1  \quad& \text{otherwise.}
            \end{cases}
\end{align}
The correct relative ordering of $\bP$ can be defined as $\bPast$
where $p^{\ast}_{ij} = 0, \forall i, j$.
The loss among the top $k$ candidates can then be written as,
\begin{align}
    \label{EQ:delta}
        \Delta (\bPast, \bP) = \frac{1}{m \cdot k} \; \sum_{i=1}^m
           \sum_{j=1}^{k} p_{i,(j)},
\end{align}
where $(j)$ denotes the index of the retrieved candidates
ranked in the $j$-th position among all top $k$ best candidates.
We define the joint feature map, $\psi$, of the form:
\begin{align}
    \label{EQ:featmap}
        \psi (\bS, \bP) = \frac{1}{m \cdot k}
            \sum_{i=1}^m \sum_{j=1}^\mprime
            (1 - p_{ij}) (\bdjm - \bdip),
\end{align}
where $\bS$ represent a set of triplets generated from
the training data,
$\bdjm$ = $[d_1(\bxi,\bxijm),$ $\cdots,$ $d_t(\bxi,\bxijm)]$ and
$\bdip$ = $[d_1(\bxi,\bxip),$ $\cdots,$ $d_t(\bxi,\bxip)]$.
The choice of $\psi(\bS, \bP)$
guarantees that the variable $\bw$, which optimizes
$\bw^\T \psi(\bS, \bP)$, will also produce
the distance function
$d(\cdot,\cdot) = \sum_{t=1}^T w_t d_t(\cdot,\cdot)$ that
achieves the optimal average recognition rate among
the top $k$ candidates.
The above problem can be summarized as the following convex
optimization problem:
\begin{align}
    \label{EQ:struct}
    \min_{ \bw , \xi }   \quad
    &
    \frac{1}{2} \| \bw  \|_{2}^{2} + \nu \, \xi    \\ \notag
    \st \; &
    \bw^\T \bigl( \psi(\bS, \bPast) - \psi(\bS, \bP) \bigr)
    \geq \Delta (\bP^{\ast}, \bP) - \xi,
\end{align}
$\forall \bP$ and $\xi \geq 0$.
Here $\bPast$ denote the correct relative ordering
and $\bP$ denote any arbitrary orderings.
Similar to \CMCtriplet, we use the cutting-plane method to solve
\eqref{EQ:struct}.

\subsubsection{Cutting-plane optimization}
In this section, we illustrate how the cutting-plane
method can be used to
solve both optimization problems: \eqref{EQ:svm2} and
\eqref{EQ:struct}.
The key idea of the cutting-plane is that
a small subset of the constraints are sufficient to find an
$\epsilon$-approximate solution to the original problem.
The cutting-plane algorithm begins with an empty initial
constraint set and iteratively adds the most violated constraint set.
At each iteration, the algorithm computes the solution
over the current working set.
The algorithm then finds the most violated constraint
and add it to the working set.
The cutting-plane algorithm continues until no constraint is violated
by more than $\epsilon$.
Since the quadratic program is of constant size, the cutting-plane method
converges in a constant number of iterations.
We present our proposed \CMCstruct in Algorithm~\ref{ALG:cutting}.

The optimization problem for finding the most violated constraint
(Algorithm~\ref{ALG:cutting}, step \ctwo) can be written as,
\begin{align}
    \label{EQ:violated1}
    \bPbar &= \max_{ \bP }  \Delta (\bP^{\ast}, \bP) -
                \bw^\T \bigl( \psi(\bS, \bPast) - \psi(\bS, \bP) \bigr) \\ \notag
           &= \max_{ \bP }  \Delta (\bPast, \bP) - \frac{1}{mk}
         \sum_{i,j}  p_{ij} \bw^\T (\bdjm - \bdip ) \\ \notag
           &= \max_{ \bP } \sum_{i=1}^m \Bigl(
                 \sum_{j=1}^k p_{i,(j)} (1-\bw^\T \bd_{i,(j)}^{\pm}  )
               - \sum_{j=k+1}^\mprime p_{i,(j)} \bw^\T \bd_{i,(j)}^{\pm} \Bigr)
\end{align}
where $\bd_{i,(j)}^{\pm} = \bdjbm - \bdip$.
Since $p_{ij}$ in \eqref{EQ:violated1} is independent,
the solution to \eqref{EQ:violated1} can be solved by maximizing over
each element $p_{ij}$.
Hence $\bPbar$ that most violates the constraint corresponds to,
\begin{align}
  \bar{p}_{i,(j)} = \left\{ \begin{array}{ll}
  \b1 \bigl( \bw^\T (\bdjbm - \bdip) \leq 1 \bigr) , &\mbox{{if   }} j \in \{ 1, \cdots, k \} \\
  \b1 \bigl( \bw^\T (\bdjbm - \bdip) \leq 0 \bigr), &\mbox{{otherwise.}}
\end{array} \right. \notag
\end{align}

For \CMCtriplet, one replaces $g(\bS, \bP, \bw)$
in Algorithm~\ref{ALG:cutting} with
$g(\bS, \bw) = 1 - \frac{1}{m(m-1)} \bw^\T \left[ \sum_{i,j} ( \bdjm - \bdip ) \right]$
and repeats the same procedure.

In this section we assume that the base metrics,
$\{d_1(\cdot,\cdot)$,$\cdots$,$d_t(\cdot,\cdot)\}$, are provided.
In the next section, we introduce two base metrics adopted in
our proposed approaches.

\SetKwInput{KwInit}{Initialize}
\SetKwRepeat{Repeat}{Repeat}{Until}

\begin{algorithm}[t]
\caption{Cutting-plane algorithm for solving coefficients of base metrics (\CMCstruct)
}
\begin{algorithmic}
\footnotesize{
   \KwIn{
     \\1)   A set of base metrics of the same individual and different individuals $\{\bdip, \bdjm\}$;
     $
     \;
     $
     \\2)    The regularization parameter, $\nu$;
     $
     \;
     $
     \\3)    The cutting-plane termination threshold, $\epsilon$;
   }

   \KwOut{
      The base metrics' coefficients $\bw$,
    }

\KwInit {
    The working set, $\cC = \varnothing;$
}

$g(\bS, \bP, \bw) = \Delta (\bPast, \bP) - \frac{1}{mk}
         \sum_{i,j}  p_{ij} \bw^\T (\bdjm - \bdip );$

\Repeat{ $ g(\bS, \bP, \bw) \leq \xi + \epsilon $
}{
\cone\ Solve the primal problem using linear SVM,
\begin{flalign}
    \notag
    \min_{ \bw , \xi } \;
    \frac{1}{2} \| \bw  \|_{2}^{2} + \nu \, \xi  \quad
    \st \;
    g(\bS, \bP, \bw) \leq \xi, \forall \bP \in \cC;
\end{flalign}
\ctwo\ Compute the most violated constraint,
\begin{align}
    \notag
    \bPbar = \max_{ \bP }  g(\bS, \bP, \bw);
\end{align}
\cthree\ $\cC \leftarrow \cC \cup  \{ \bPbar \};$
}
} %
\end{algorithmic}
\label{ALG:cutting}
\end{algorithm}

\subsection{Base metrics}
\label{sec:metric}
Metric learning can be divided into two categories:
linear \cite{Davis2007Information,Kostinger2012Large,Weinberger2006Distance}
and non-linear methods \cite{Chopra2005Learning,Frome2007Learning,
Kedem2012Nonlinear,Weinberger2008Fast,Xiong2014Person}.
In the linear case, the goal is to learn a linear mapping
by estimating a matrix $\bM$
such that the distance between images of the same individual,
$(\bxi - \bxip)^\T \bM (\bxi - \bxip)$, is less than
the distance between images of different individuals,
$(\bxi - \bxijm)^\T \bM (\bxi - \bxijm)$.
The linear method can be easily extended to learn non-linear
mapping by kernelization \cite{Shawe2004Kernel}.
The basic idea is to learn a linear mapping in the feature space of some
non-linear function, $\phi$,
such that
the distance
$( \phi(\bxi) - \phi(\bxip) )^\T \bM ( \phi(\bxi) - \phi(\bxip) )$
is less than
the distance
$( \phi(\bxi) - \phi(\bxijm) )^\T \bM ( \phi(\bxi) - \phi(\bxijm) )$.

\paragraph{Metric learning from equivalence constraints}
The basic idea of KISS metric learning
(KISS ML) \cite{Kostinger2012Large}, is to
learn the Mahalanobis distance by considering a log likelihood ratio
test of two Gaussian distributions.
The likelihood ratio test between dissimilar pairs and similar pairs
can be written as,
\begin{align}
    \label{EQ:KISSME1}
    r(\bxi,\bxj) = \log \frac{ \frac{1}{c_d} \exp(-\frac{1}{2} \bx_{ij}^\T \Sigma_{\cD}^{-1} \bx_{ij}  )}
                             { \frac{1}{c_s} \exp(-\frac{1}{2} \bx_{ij}^\T \Sigma_{\cS}^{-1} \bx_{ij}  )} ,
\end{align}
where $\bx_{ij} = \bx_{i} - \bx_{j}$,
$c_d = \sqrt{2 \pi | \Sigma_{\cD} |} $,
$c_s = \sqrt{2 \pi | \Sigma_{\cS} |} $,
$\Sigma_{\cD}$ and $\Sigma_{\cS}$ are covariance matrices of
dissimilar pairs and similar pairs, respectively.
By taking log and discarding constant terms,
\eqref{EQ:KISSME1} can be simplified as,
\begin{align}
    \label{EQ:KISSME2}
    r(\bxi,\bxj) = (\bx_{i} - \bx_{j})^\T (\Sigma_{\cS}^{-1} - \Sigma_{\cD}^{-1})
                      (\bx_{i} - \bx_{j}),
\end{align}
Hence the Mahalanobis distance matrix $\bM$ can be written as
$\Sigma_{\cS}^{-1} - \Sigma_{\cD}^{-1}$.
The authors of \cite{Kostinger2012Large} clip the spectrum of $\bM$ by eigen-analysis
to ensure $\bM$ is positive semi-definite.
This simple algorithm has shown to perform surprisingly well on
the person re-id problem
\cite{Roth2014Mahalanobis, Li2014Deep}.

\paragraph{Kernel-based metric learning}
There exist several non-linear extensions to metric learning.
In this section, we introduce recently proposed kernel-based
metric learning, known as kernel Local Fisher Discriminant
Analysis (kLFDA) \cite{Xiong2014Person},
which is a non-linear extension to the previously
proposed LFDA \cite{Pedagadi2013Local} and
has demonstrated the state-of-the-art
performance on iLIDS, CAVIAR and 3DPeS data sets.
The basic idea of kLFDA is to find a projection matrix $\bM$
which maximizes the between-class scatter matrix while minimizing
the within-class scatter matrix using the Fisher discriminant
objective.
Similar to LFDA, the projection matrix can be estimated using
generalized Eigenvalues.
Unlike LFDA, kLFDA represent the projection matrix with the
data samples in the kernel space $\phi(\cdot)$.
\subsection{Visual features}
\label{subsec:feat}
We introduce visual features which have been applied in
our person re-id approaches.

\paragraph{SIFT$/$LAB patterns}
Scale-invariant feature transform (SIFT) has
gained a lot of research attention
due to its invariance to scaling, orientation and illumination
changes \cite{Lowe2004Distinctive}.
The descriptor represents occurrences of gradient orientation
in each region.
In this work, we combine discriminative SIFT with color histogram
extracted from the LAB colorspace.

\paragraph{LBP$/$RGB patterns}
Local Binary Pattern (LBP) is another feature descriptor
that has received a lot of attention in the literature
due to its effectiveness and efficiency \cite{Ojala2002Multiresolution}.
The standard version of $8$-neighbours LBP has a radius of $1$
and is formed by thresholding the $3 \times 3$
neighbourhood with the centre pixel's value.
To improve the classification accuracy of LBP,
we combine LBP histograms with color histograms extracted
from the RGB colorspace.

\paragraph{Region covariance patterns}
Region covariance is another texture descriptor which has
shown promising results in texture classification \cite{Tuzel2006Region}.
The covariance descriptor is extracted from the covariance
of several image statistics inside a region of interest \cite{Tuzel2006Region}.
Covariance matrix provides a measure of the relationship between
two or more set of variates.
The diagonal entries of covariance matrices represent the variance
and the non-diagonal entries represent the correlation value
between low-level features.

\paragraph{Neural patterns}
Large amount of available training data and
increasing computing power have lead to
a recent success of deep
convolutional neural networks (CNN)
on a large number of computer vision applications.
CNN exploits the strong spatially local correlation present in
natural images by enforcing a local connectivity pattern
between neurons of adjacent layers.
In the deep CNN architecture, convolutional layers are placed
alternatively between max-pooling and contrast
normalization layers \cite{Krizhevsky2012Imagenet}.

\paragraph{Implementation}
See supplementary for detailed implementation.

\section{Experiments}
\paragraph{Datasets}
There exist several challenging benchmark data sets for person re-identification.
In this experiment, we select four commonly used data sets
(iLIDS, 3DPES, PRID2011, VIPeR)
and two recently introduced data sets with a large number of individuals
(CUHK01 and CUHK03).
The iLIDS data set has $119$ individuals captured from eight cameras with
different viewpoints \cite{Zheng2009Associating}.
The number of images for each individual
varies from $2$ to $8$, \ie, eight cameras are used to capture $119$ individuals.
The data set consists of large occlusions caused by people and luggages.
The 3DPeS data set is designed mainly for people tracking
and person re-identification \cite{Baltieri20113dpes}.
It contains numerous video sequences taken from a real surveillance environment
with eight different surveillance cameras and
consists of $192$ individuals.
The number of images for each individual varies from $2$ to $26$ images.
The Person RE-ID 2011 (PRID2011) data set consists of images extracted
from multiple person trajectories recorded from two surveillance
static cameras \cite{Hirzer2011Person}.
Camera view A contains $385$ individuals, camera view B contains $749$ individuals,
with $200$ of them appearing in both views.
Hence, there are $200$ person image pairs in the dataset.

VIPeR is one of the most popular used data sets for
person re-identification \cite{Gray2007Evaluating}.
It conntains $632$ individuals taken from two cameras with
arbitrary viewpoints and varying illumination conditions.
The CUHK01 data set contains $971$ persons captured from two camera views in a campus
environment \cite{Li2012Human}.
Camera view A captures the frontal or back view of the individuals while
camera view B captures the profile view.
Finally, the CUHK03 data set consists of $1360$ persons taken from six cameras \cite{Li2014Deep}
The data set consists of manually cropped pedestrian images and
images cropped from the pedestrian detector of \cite{Felzenszwalb2010Object}.
Due to the imperfection in the pedestrian detector, which
causes some misalignments of cropped images,
we use images which are manually annotated by hand.

\begin{figure*}[t]
    \centering
        \includegraphics[width=0.3\textwidth,clip]{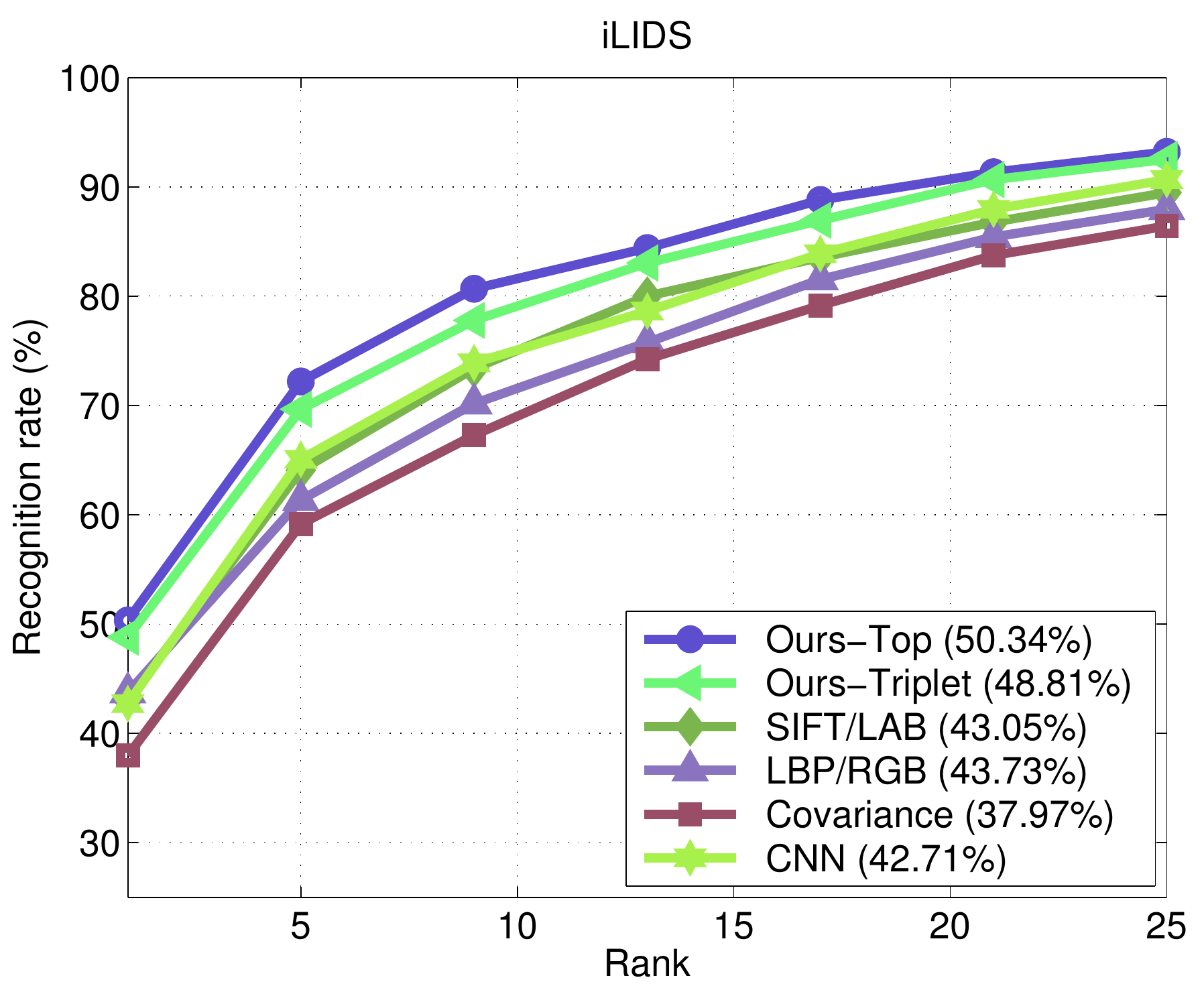}
        \includegraphics[width=0.3\textwidth,clip]{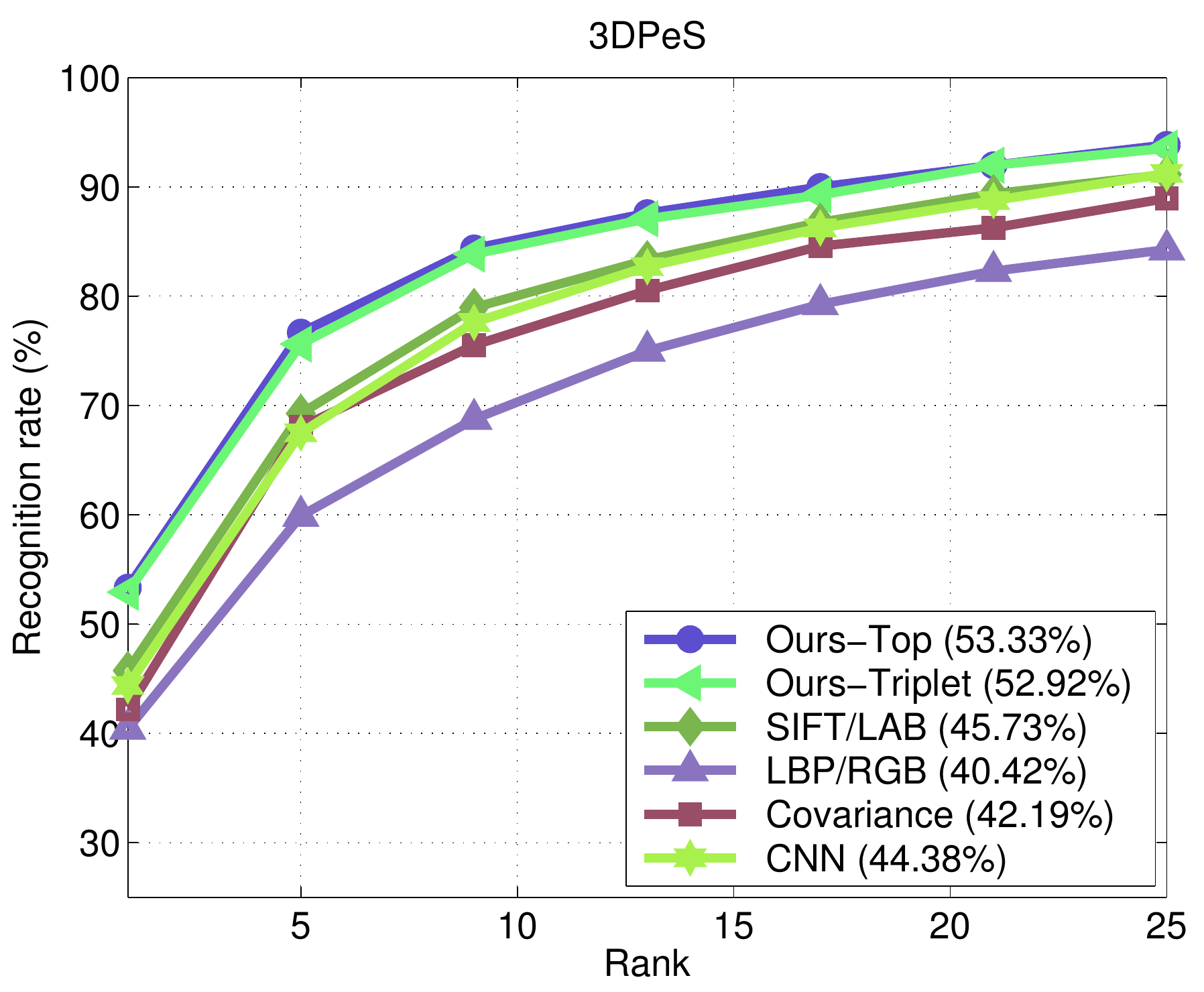}
        \includegraphics[width=0.3\textwidth,clip]{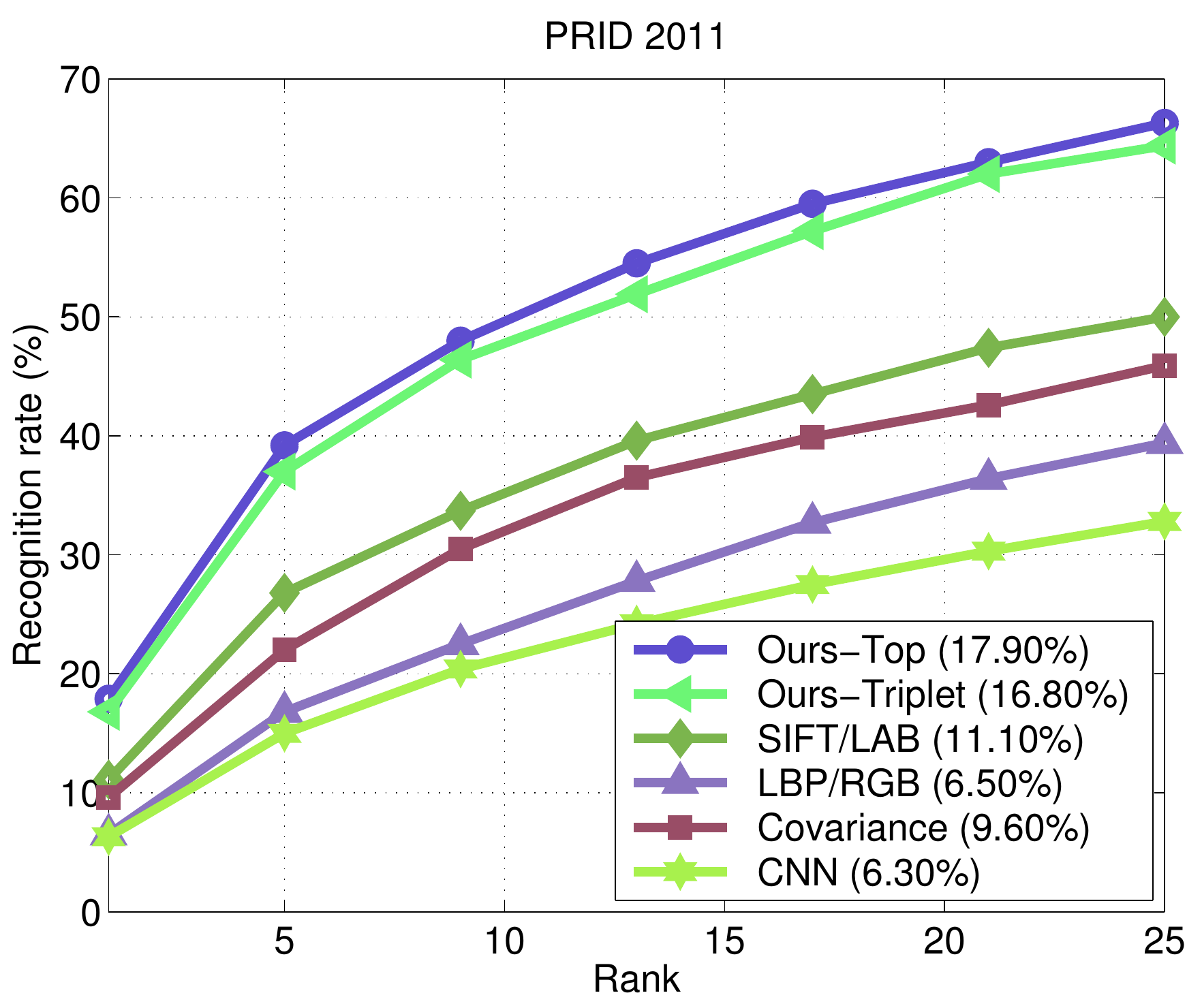}
        \includegraphics[width=0.3\textwidth,clip]{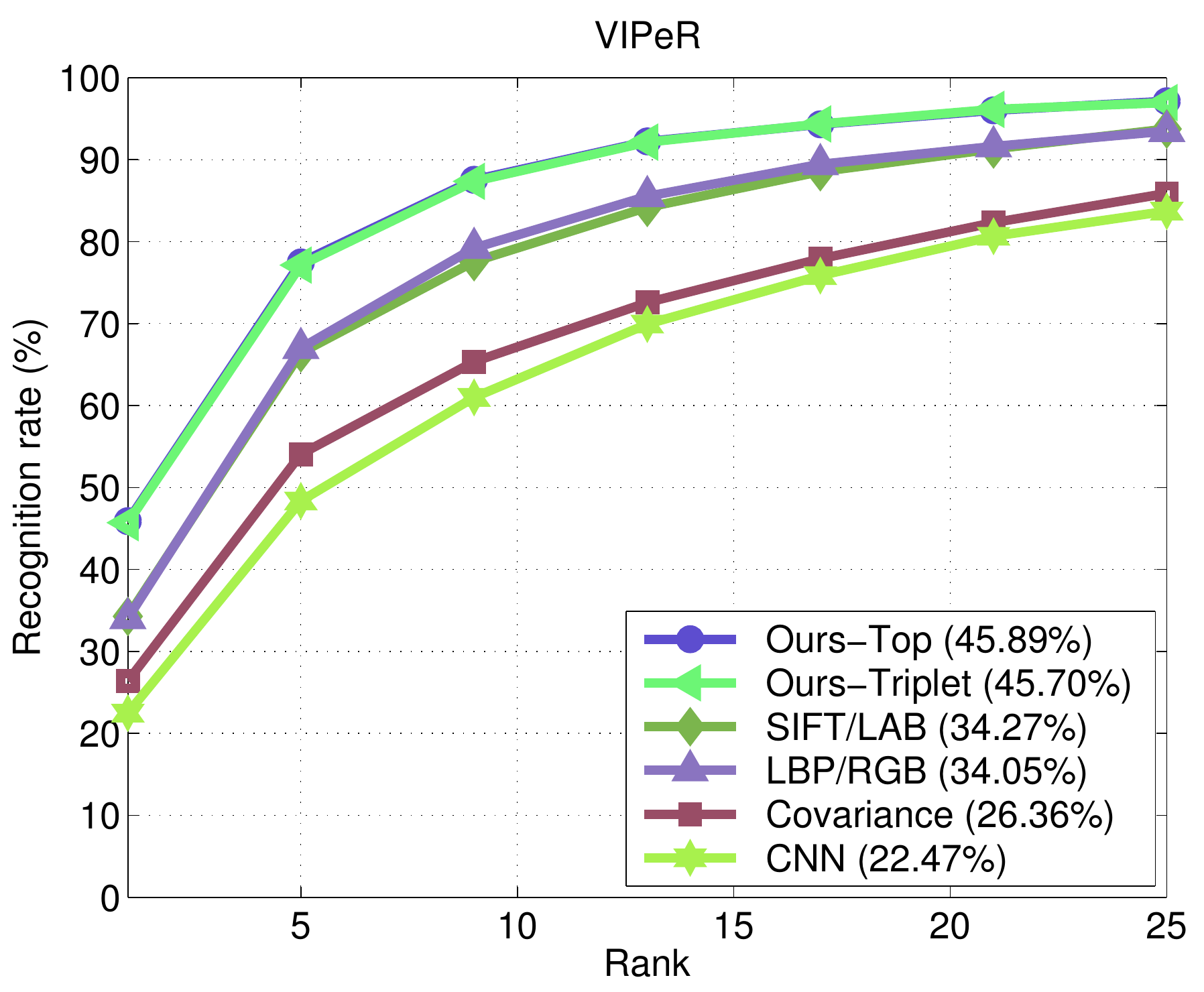}
        \includegraphics[width=0.3\textwidth,clip]{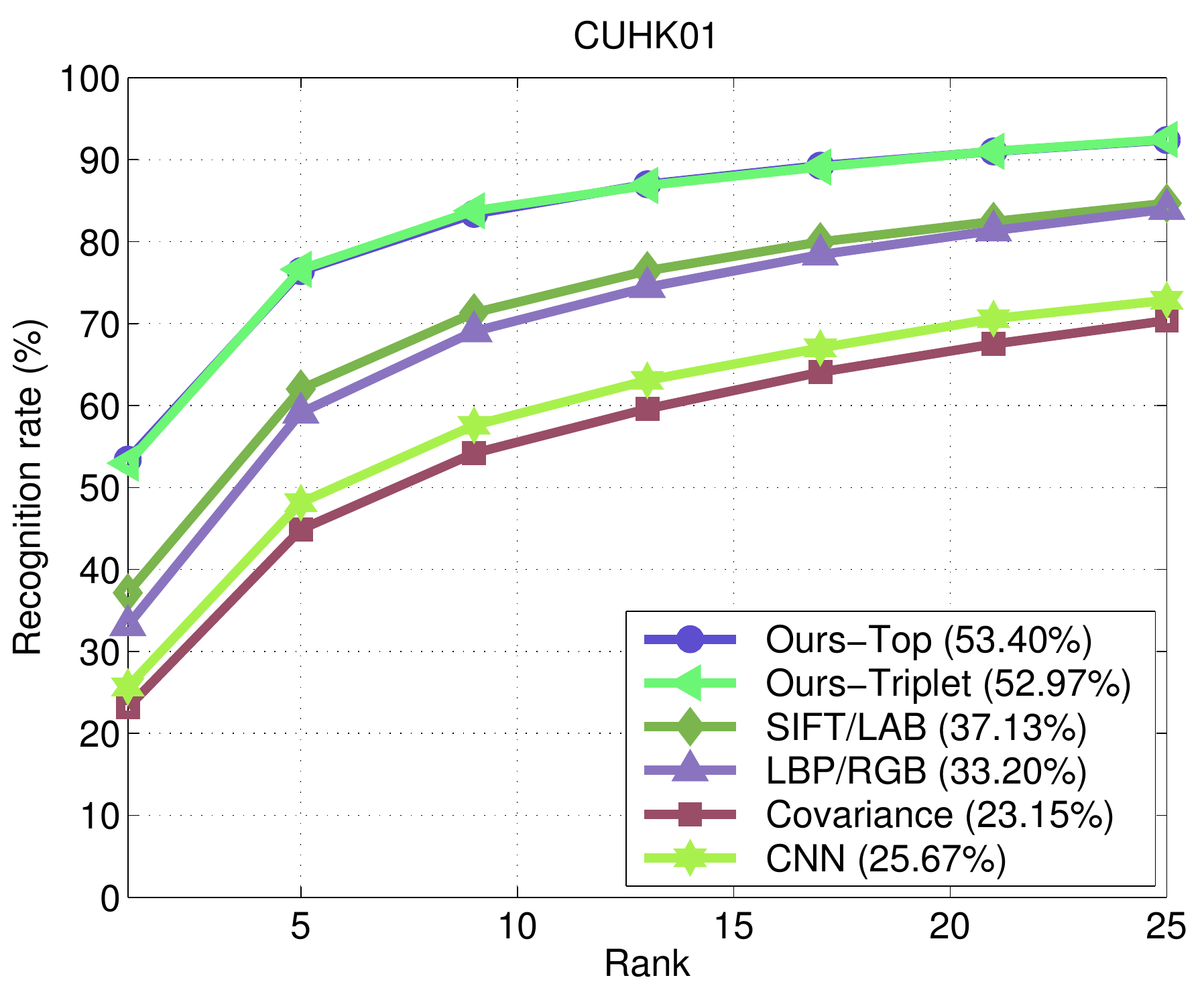}
        \includegraphics[width=0.3\textwidth,clip]{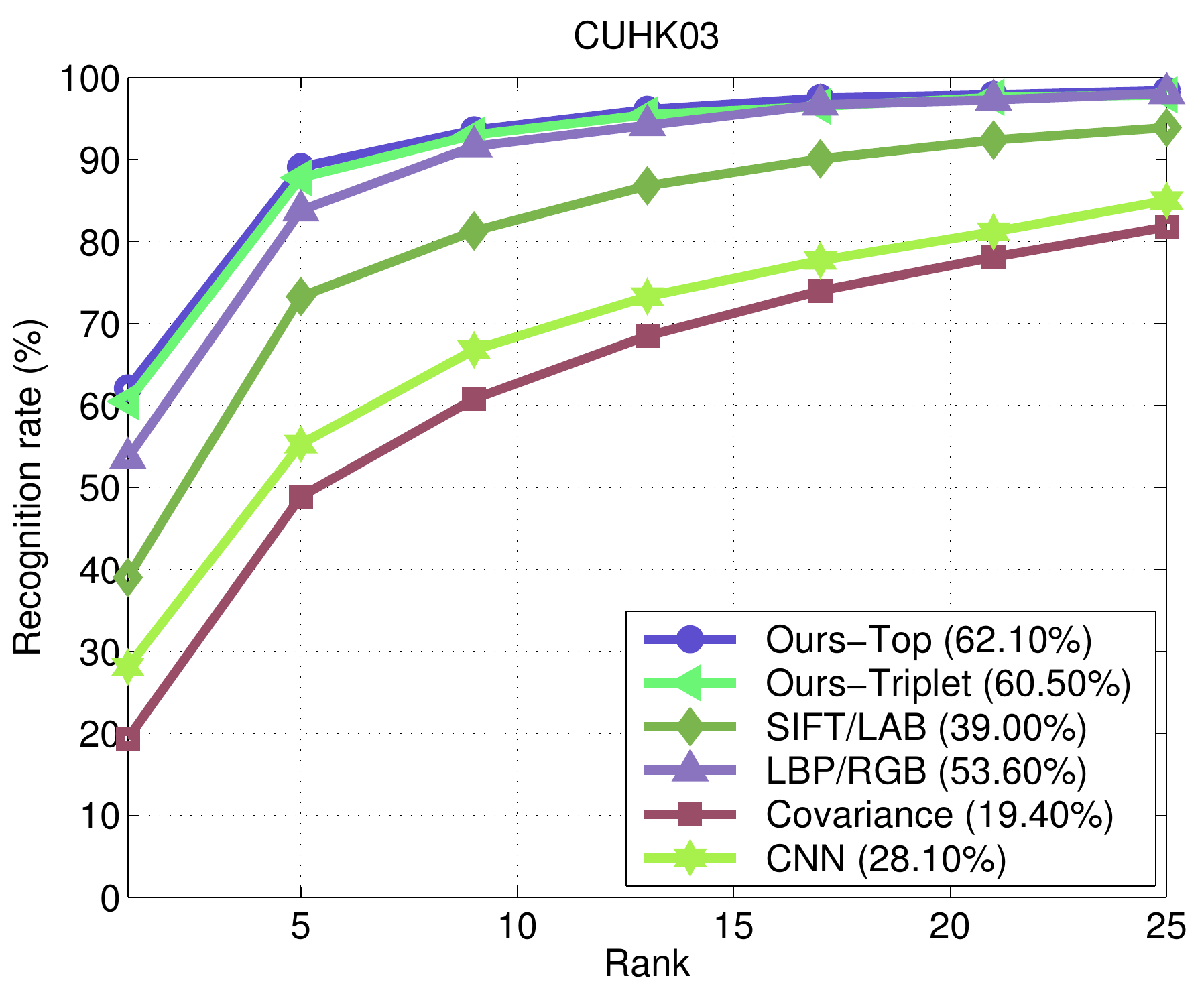}
    \caption{
    Performance comparison of base metrics with
    different visual features:  SIFT$/$LAB, LBP$/$RGB,
    covariance descriptor and CNN features.
    Rank-$1$ recognition rates are shown in parentheses.
    The higher the recognition rate, the better the performance.
    \textbf{Ours-Top} (\CMCstruct) represents our ensemble approach which optimizes the CMC score
    over the top $k$ returned candidates.
    \textbf{Ours-Triplet} (\CMCtriplet) represents our ensemble approach which
    minimizes the number of returned candidates such that the rank-$k$ recognition
    rate is equal to one.
    }
    \label{fig:feat1}
\end{figure*}

\begin{figure*}[t]
    \centering
        \includegraphics[width=0.3\textwidth,clip]{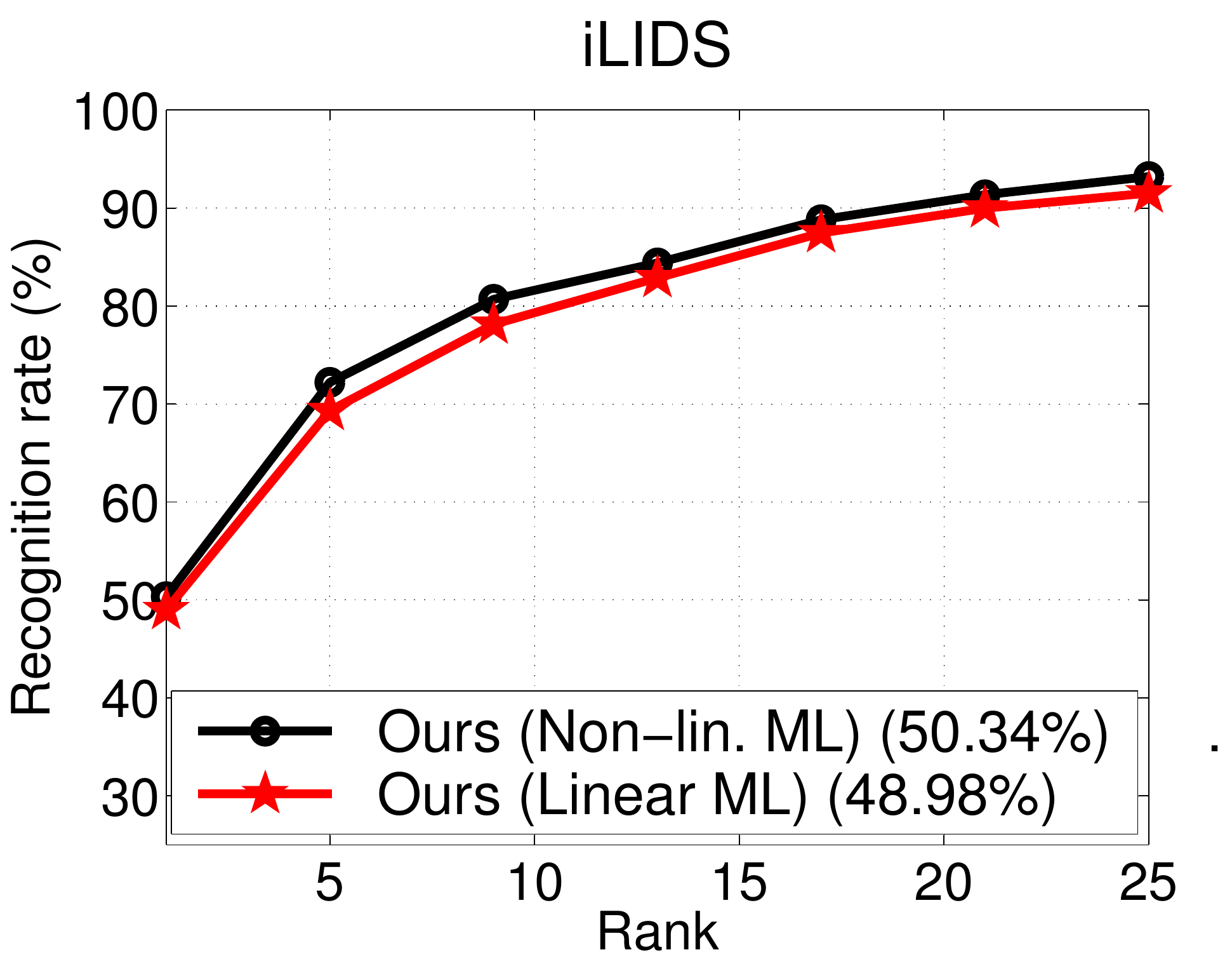}
        \includegraphics[width=0.3\textwidth,clip]{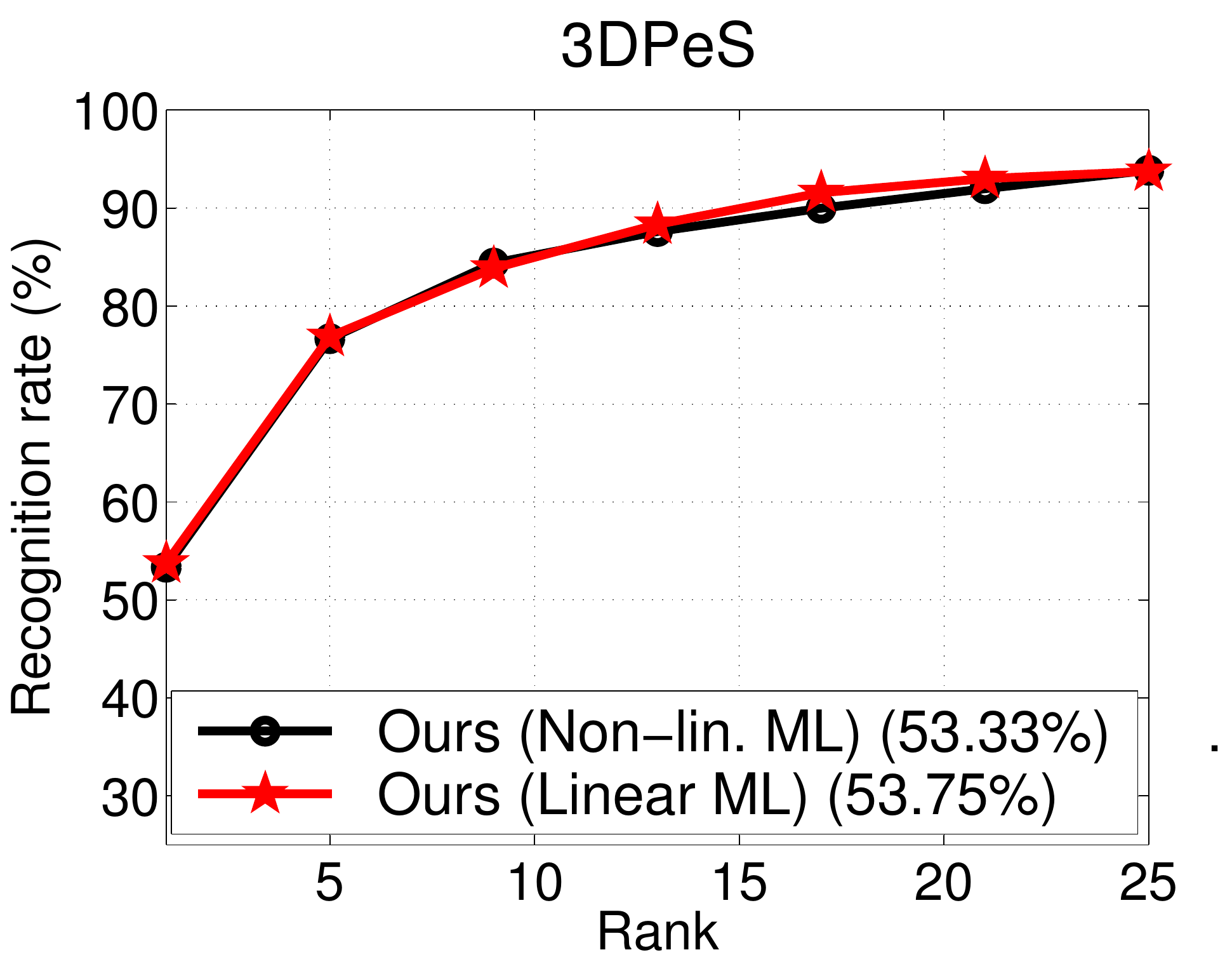}
        \includegraphics[width=0.3\textwidth,clip]{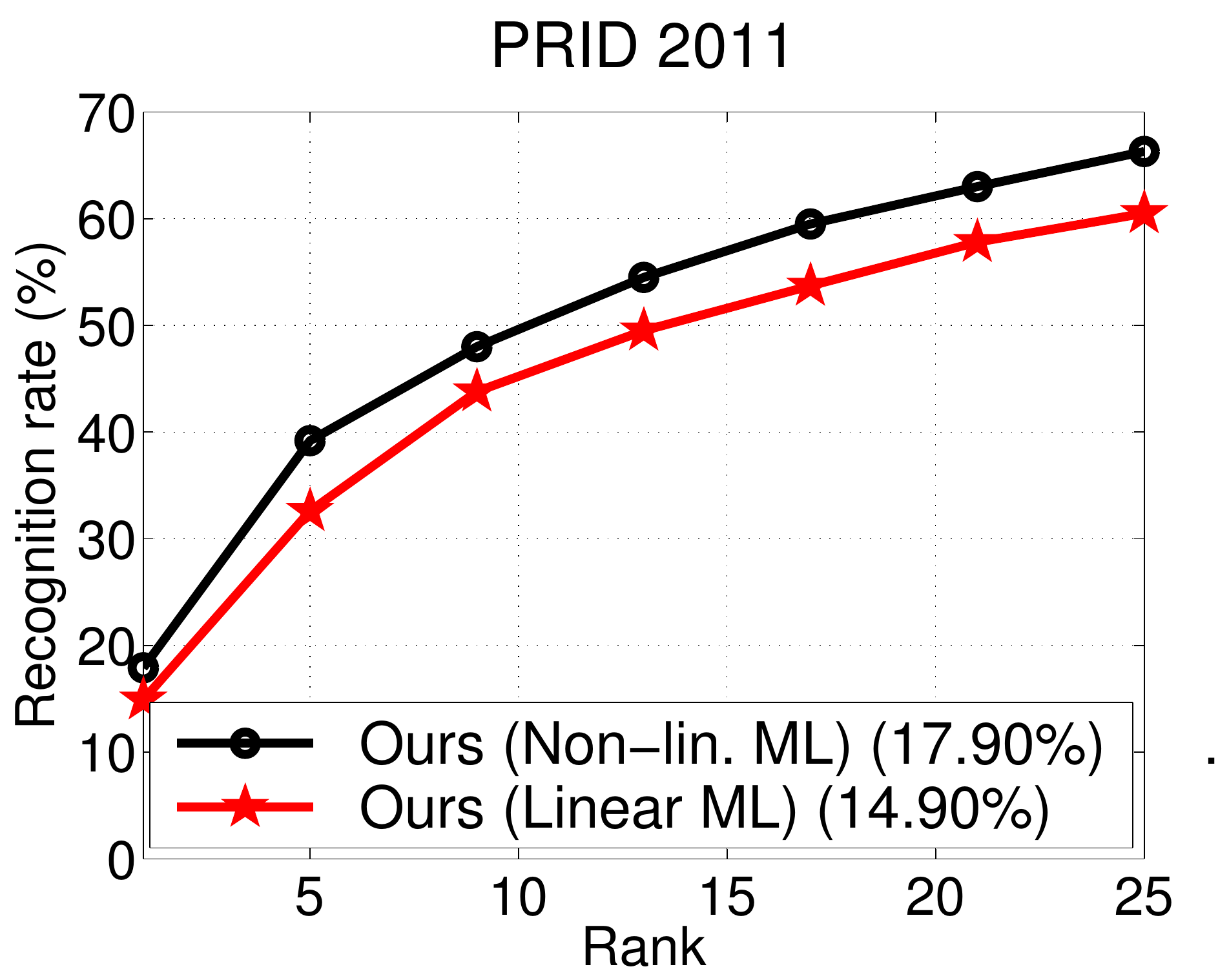}
        \includegraphics[width=0.3\textwidth,clip]{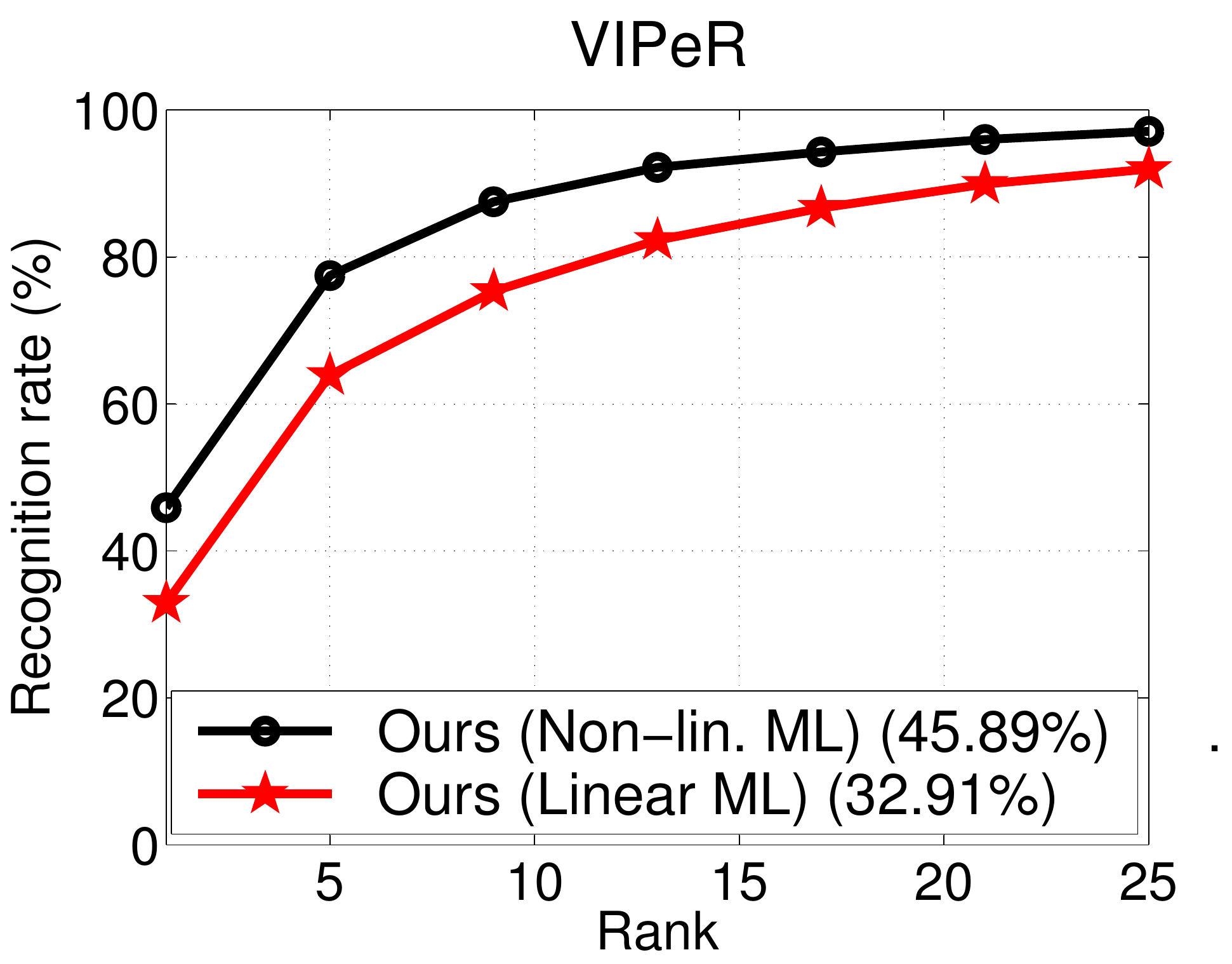}
        \includegraphics[width=0.3\textwidth,clip]{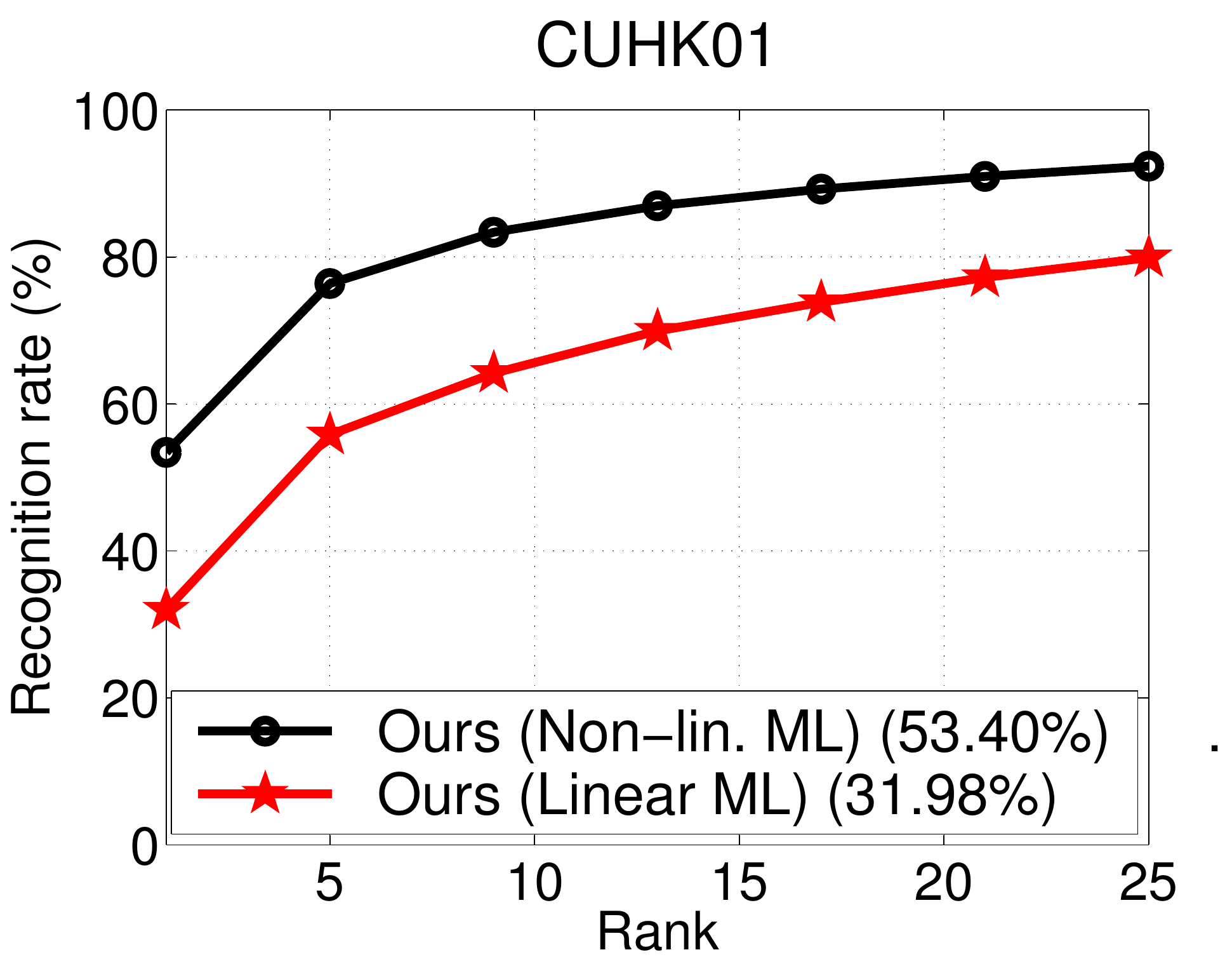}
        \includegraphics[width=0.3\textwidth,clip]{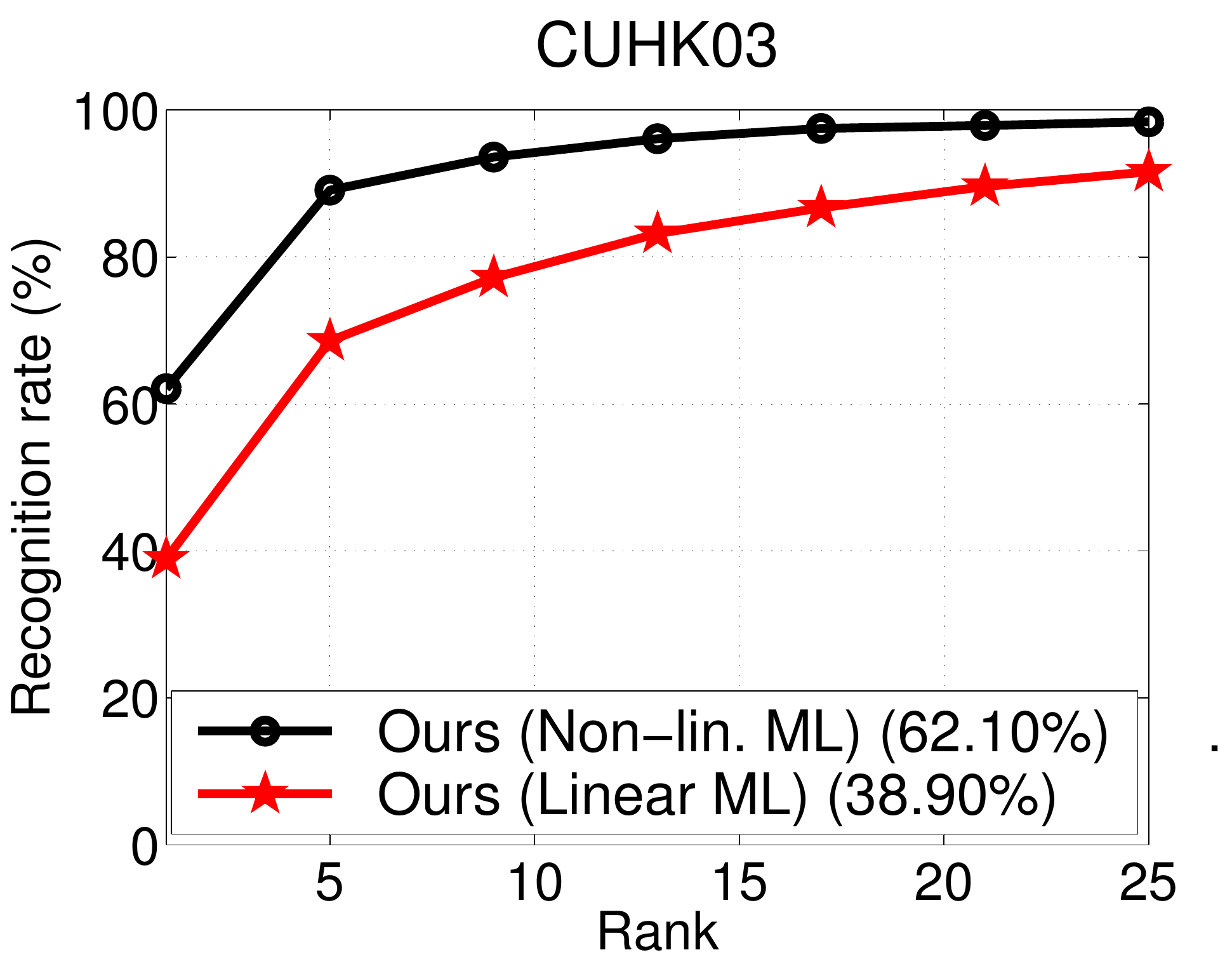}
    \caption{
    Performance comparison of \CMCstruct with two different base metrics:
    linear base metric (Linear Metric Learning) and
    non-linear base metric (Non-lin. Metric Learning).
    On VIPeR, CUHK01 and CUHK03 data sets,
    an ensemble of non-linear base metrics significantly outperforms
    an ensemble of linear base metrics.
    }
    \label{fig:linear_nonlin}
\end{figure*}

\paragraph{Evaluation protocol}
In this paper, we adopt a single-shot experiment
setting, similar to \cite{Li2013Learning,
Pedagadi2013Local, Xiong2014Person, Zhao2014Learning, Zheng2011Person}.
For all data sets except CUHK03, all the individuals in the data set are
randomly divided into two subsets so that
the training set and the test
set contains half of the available individuals with no overlap on person identities.
For data set with two cameras, we randomly select one image of the individual
taken from camera view A as the probe image
and one image of the same individual taken from camera view B
as the gallery image.
For multi-camera data sets, two images of the same individual
are chosen: one is used as the probe image and the other as the gallery image.
For CUHK03, we set the number of individuals in the train$/$test split
to $1260$$/$$100$ as conducted in \cite{Li2014Deep}.
To be more specific, there are $59$, $96$, $100$, $316$, $485$ and $100$ individuals
in each of the test split for the iLIDS, 3DPeS, PRID2011, VIPeR, CUHK01 and CUHK03
data sets, respectively.
The number of probe images (test phase) is equal to the number of
gallery images in all data sets except
PRID2011, in which the number of probe images is $100$
and the number of test gallery images is $649$
(all images from camera view B except the $100$ training samples).
This procedure is repeated $10$ times and the
average of cumulative matching charateristic (CMC) curves across
$10$ partitions is reported.
The CMC curve provides a ranking for every image in the gallery
with respect to the probe.

\paragraph{Parameters setting}
For the linear base metric (KISS ML \cite{Kostinger2012Large}),
we apply principal component analysis (PCA) to
reduce the dimensionality and remove noise.
Without performing PCA, it is computationally infeasible to
inverse covariance matrices of both similar and dissimilar
pairs as discussed in \cite{Kostinger2012Large}.
For each visual feature, we
reduce the feature dimension to $64$ dimensional subspaces.
For the non-linear base metric (kLFDA \cite{Xiong2014Person}),
we set the regularization parameter
for class scatter matrix to $0.01$,
\ie, we add a small identity matrix to the class scatter matrix.
For both SIFT$/$LAB and LBP$/$RGB features, we apply the RBF-$\chi^2$ kernel.
For region covariance and CNN features, we apply
the Gaussian RBF kernel $\kappa(\bx,\bx^\prime) = \exp(-\| \bx - \bx^\prime \| / \sigma^2)$.
The kernel parameter is tuned to an appropriate value for each
data set.
In this experiment, we set the value of $\sigma^2$ to be the
same as the first quantile of all distances \cite{Xiong2014Person}.

For \CMCtriplet, we
choose the regularization parameter
($\nu$ in \eqref{EQ:svm}) from
$\{10^{3}$,$10^{3.1}$,$\cdots$,$10^{4}\}$ by cross-validation
on the training data.
For \CMCstruct, we choose the
regularization parameter ($\nu$ in \eqref{EQ:struct})
from $\{10^{2}$,$10^{2.1}$,$\cdots$,$10^{3}\}$ by cross-validation
on the training data.
We set the cutting-plane termination threshold to $10^{-6}$.
The recall parameter ($k$ in \eqref{EQ:featmap}) is set
to be $10$ for iLIDS, 3DPeS, PRID2011
and VIPeR and $40$ for larger data sets (CUHK01 and CUHK03).
Since the success of metric learning algorithms often depends on the choice
of good parameters, we train multiple metric learning for each feature.
Specifically, for KISS ML, we reduce their feature dimensionality
to $32$, $48$ and $64$ dimensions and use all three to learn
the weight $\bw$ for \CMCtriplet and \CMCstruct.
Similarly, for kFLDA, we set the
$\sigma^2$ to be the same as the \nth{5}, the
\nth{10} and the first quantile of all distances.

\subsection{Evaluation and analysis}
\paragraph{Feature evaluation}
We investigate the impact of low-level and high-level
visual features on the recognition performance of
person re-identification.
Fig.~\ref{fig:feat1} shows the CMC performance of different
visual features and their rank-$1$ recognition rates
when trained with the
kernel-based LFDA
(non-linear metric learning) on six benchmark
data sets.
On VIPeR, CUHK01 and CUHK03 data sets,
we observe that both SIFT$/$LAB and LBP$/$RGB
significantly outperform
covariance descriptor and CNN features.
This result is not surprising since
SIFT$/$LAB combines edges and color features
while LPB$/$RGB combines texture and color features.
We suspect the use of color
helps improve the overall recognition performance
of both features.
We observe that CNN features perform poorer than
hand-crafted low-level features in our experiments.
We suspect that the CNN pre-trained model
has been designed for ImageNet object categories
\cite{Krizhevsky2012Imagenet},
in which color information might be less important.
However on many person re-id data sets,
a large number of persons wear similar types
of clothing, \eg, t-shirt and jeans,
but with different color.
Therefore color information becomes an important cue for recognizing two
different individuals.
Overall, we observe that SIFT$/$LAB features perform
well consistently on all data sets evaluated.

\paragraph{Ensemble approach with different base metrics}
Next we compare the performance of our approach
with two different base metrics: linear metric learning \cite{Kostinger2012Large}
and non-linear metric learning \cite{Xiong2014Person}
(introduced in Sec.~\ref{sec:metric}).
In this experiment, we use \CMCstruct to learn an ensemble.
Experimental results are shown in Fig.~\ref{fig:linear_nonlin}.
Two observations can be made from the figure:
1) Both approaches perform similarly when the number of train$/$test individuals
is small, \eg, on iLIDS and 3DPeS data sets;
2) Non-linear base metrics outperforms linear base metric
when the number of individuals increase.
We suspect that there is less diversity when the number
of individuals is small.
No further improvement is observed when we replace linear base
metrics with non-linear base metrics.

\begin{table*}[tb]
  \centering
  {
  \begin{tabular}{l||c|c|c||c|c|c||c|c|c}
  \hline
  \multirow{2}{*}{Rank} & \multicolumn{3}{|c||}{VIPeR} &
               \multicolumn{3}{|c||}{CUHK01} &  \multicolumn{3}{|c}{CUHK03} \\
  \cline{2-10}
            & Avg. & \CMCtriplet & \CMCstruct &
                Avg. & \CMCtriplet & \CMCstruct & Avg. & \CMCtriplet & \CMCstruct   \\
  \hline
  \hline
$1$ & $44.9$ & $45.7$ & $\mathbf{45.9}$ & $51.9$ & $53.0$ & $\mathbf{53.4}$ & $57.4$   & $60.5$   & $\mathbf{62.1}$ \\
$2$ & $58.3$ & $59.6$ & $\mathbf{60.2}$ & $63.3$ & $64.1$ & $\mathbf{64.3}$ & $71.7$   & $73.5$   & $\mathbf{76.6}$ \\
$5$ & $76.3$ & $77.1$ & $\mathbf{77.5}$ & $75.1$ & $76.1$ & $\mathbf{76.4}$ & $85.9$   & $87.8$   & $\mathbf{89.1}$ \\
$10$ & $88.2$ & $\mathbf{88.9}$ & $\mathbf{88.9}$ & $83.0$ & $ 84.0$ & $\mathbf{84.4}$ & $93.1$   & $93.5$   & $\mathbf{94.3}$ \\
$20$ & $94.9$ & $95.7$ & $\mathbf{95.8}$ & $89.4$ & $\mathbf{90.7}$ & $90.5$ & $96.9$   & $97.4$   & $\mathbf{97.8}$ \\
$50$ & $99.4$ & $\mathbf{99.5}$ & $\mathbf{99.5}$ & $95.9$ & $\mathbf{96.4}$ & $\mathbf{96.4}$ & $99.5$   & $\mathbf{99.7}$  & $\mathbf{99.7}$ \\
$100$ & $99.9$ & $\mathbf{100.0}$ & $\mathbf{100.0}$ & $\mathbf{98.6}$ & $\mathbf{98.6}$ & $\mathbf{98.6}$ & $\mathbf{100.0}$  & $\mathbf{100.0}$  & $\mathbf{100.0}$ \\

  \hline
  \end{tabular}
  }
  \caption{
  Re-id recognition rate ($\%$) at different recall (rank).
  The best result is shown in boldface.
  Both \CMCstruct and \CMCtriplet achieve similar performance
  when retrieving $\geq 50$ candidates.
  }
  \label{tab:cmc}
\end{table*}

\begin{figure}[t]
    \centering
        \includegraphics[width=0.39\textwidth,clip]{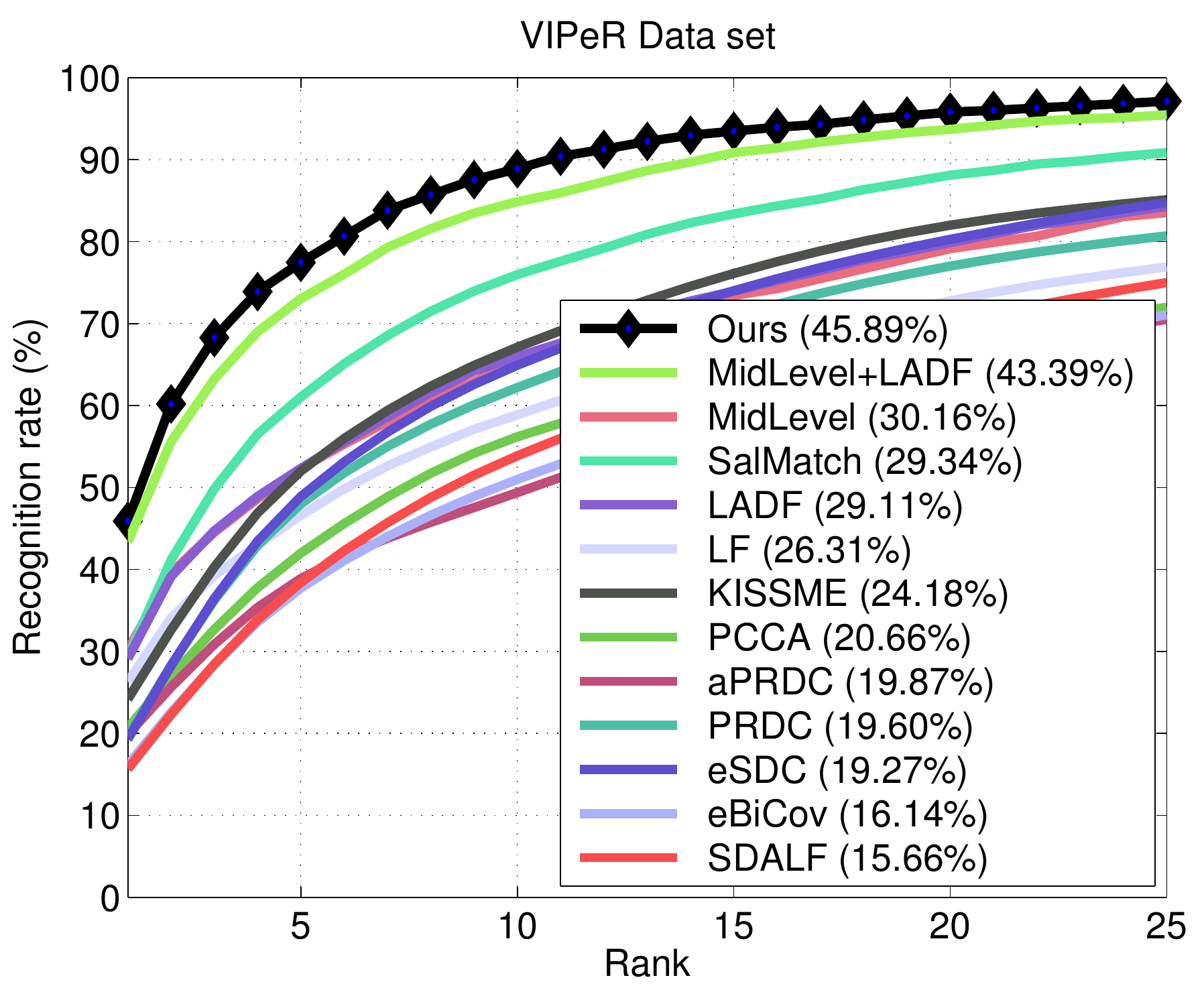}
        \includegraphics[width=0.39\textwidth,clip]{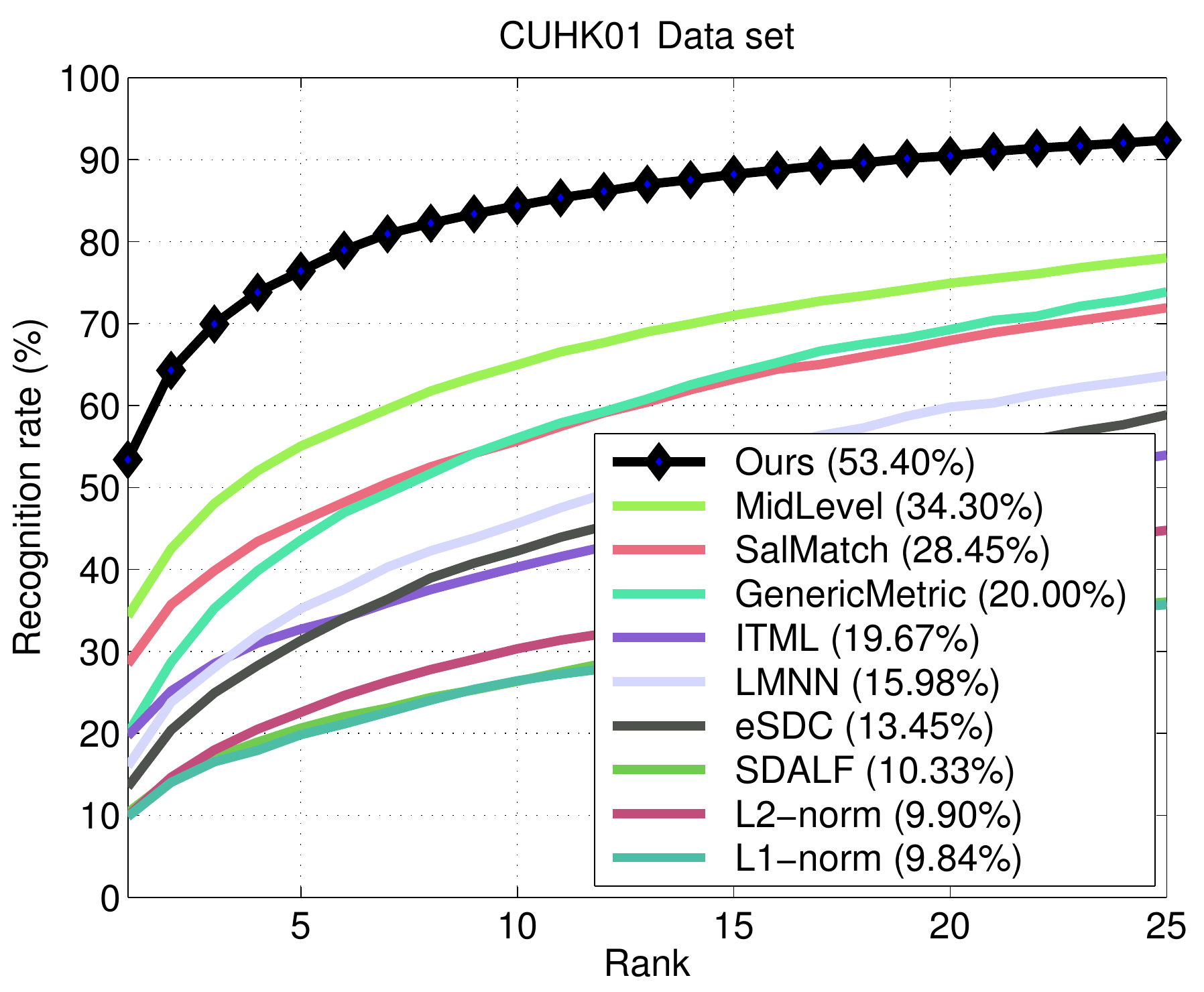}
    \caption{
    CMC performance for VIPeR and CUHK01 data sets.
    The higher the recognition rate, the better the performance.
    Our approach outperforms all existing person re-id algorithms.
    }
    \label{fig:best1}
\end{figure}

\begin{table}[bt]
  \centering
  {
  \begin{tabular}{l|c|c|c|c}
  \hline
  \multirow{2}{*}{Data set} & \multicolumn{2}{|c|}{\# Individuals} &
                    \multirow{2}{*}{Prev. best} & \multirow{2}{*}{Ours} \\
  \cline{2-3}
            & train & test &  &  \\
  \hline
  \hline
   iLIDS    &  $59$  & $60$  & $40.3\%$ \cite{Xiong2014Person} & $\mathbf{50.3\%}$  \\
   3DPeS    &  $96$  & $96$  & $\mathbf{54.2\%}$ \cite{Xiong2014Person} & $53.3\%$  \\
   PRID2011 & $100$  & $100$ & $16.0\%$ \cite{Roth2014Mahalanobis} & $\mathbf{17.9\%}$  \\
   VIPeR    & $316$  & $316$ & $43.4\%$ \cite{Zhao2014Learning} & $\mathbf{45.9\%}$ \\
   CUHK01   & $486$  & $485$ & $34.3\%$ \cite{Zhao2014Learning} & $\mathbf{53.4\%}$ \\
   CUHK03   & $1260$ & $100$ & $20.7\%$ \cite{Li2014Deep} & $\mathbf{62.1\%}$ \\
  \hline
  \end{tabular}
  }
  \caption{
  Rank-$1$ recognition rate of existing best reported results and our results.
  The best result is shown in boldface.
  }
  \label{tab:best2}
\end{table}

\paragraph{Performance at different recall values}
Next we compare the performance of the proposed \CMCtriplet
with \CMCstruct.
Both optimization algorithms optimize the recognition
rate of person re-id but with different objective criteria.
We compare the performance of both algorithms
with the baseline approach, in which we simply set the value of
$\bw$ to a uniform weight.
Since distance functions of different features have different scales,
we normalize the distance between each probe image to all images in
the gallery to be between zero and one.
In other words,
we set the distance between the probe image and
the nearest gallery image
to be zero and the distance between
the probe image and the furthest gallery
image to be one.
The matching accuracy is shown in Table~\ref{tab:cmc}.
We observe that \CMCstruct achieves the best recognition rate performance
at a small recall value.
At a large recall value (rank $\geq 50$), both \CMCstruct and \CMCtriplet
perform similarly.
Interestingly, a simple averaging performs quite well on VIPeR, in
which the number of individuals in the test set is small.

\subsection{Comparison with state-of-the-art results}

Fig.~\ref{fig:best1} compares our results with
other person re-id algorithms on two
major benchmark data sets: VIPeR and CUHK01.
Our approach outperforms all existing person re-id algorithms.
Next we compare our results with
the best reported results in the literature.
The algorithm proposed in \cite{Xiong2014Person} achieves
state-of-the-art results on iLIDS and 3DPeS
data sets ($40.3\%$ and $54.2\%$ recognition rate at rank-$1$, respectively).
Our approach outperforms \cite{Xiong2014Person} on the iLIDS
($50.3\%$) and
achieve a comparable result on 3DPeS ($53.3\%$).
Zhao \etal propose mid-level
filters for person re-identification \cite{Zhao2014Learning},
which achieve state-of-the-art results on the VIPeR and CUHK01 data sets
($43.39\%$ and $34.30\%$ recognition rate at rank-$1$, respectively).
Our approach outperforms \cite{Zhao2014Learning} by
achieving a recognition rate of $45.89\%$ and $53.40\%$ on the
VIPeR and CUHK01 data sets, respectively.
Table~\ref{tab:best2} compares our results
with other state-of-the-art methods on other person
re-identification data sets.

\section{Conclusion}
In this paper, we present an effective structured learning based
approach for person
re-id by combining multiple low-level and high-level
visual features into a single framework.
Our approach is practical to real-world applications
since the performance can be concentrated in the range of
most practical importance.
Moreover our proposed approach is  flexible and can be applied
to any metric learning algorithms.
Experimental results demonstrate the effectiveness of the
proposed approach on six major person re-id data sets.

{\small
\bibliographystyle{abbrv}
\bibliography{draft}
}

\end{document}